\documentclass[conference]{IEEEtran}
\IEEEoverridecommandlockouts
\usepackage{cite}
\usepackage{amsmath,amssymb,amsfonts}
\usepackage{algorithmic}
\usepackage{graphicx}
\usepackage{textcomp}
\usepackage{cite}
\usepackage{hyperref}
\usepackage{booktabs}
\usepackage[switch]{lineno} 
\usepackage{xcolor}
\def\BibTeX{{\rm B\kern-.05em{\sc i\kern-.025em b}\kern-.08em
    T\kern-.1667em\lower.7ex\hbox{E}\kern-.125emX}}

\newcommand{\shortname}{HFNeRF}
\newcommand{\shortnamewithspace}{HFNeRF }
\newcommand{\longname}{HFNeRF: Learning Human Biomechanic Features with Neural Radiance Fields}
\begin{document}

\title{\longname\\
}
\author{Arnab Dey\textsuperscript{1} \\
{\tt\small adey@i3s.unice.fr}
  \and
  Di Yang\textsuperscript{2}\\
{\tt\small di.yang@inria.fr}
  \and
  Antitza Dantcheva \textsuperscript{2}\\
  {\tt\small antitza.dantcheva@inria.fr}
  \and
  Jean Martinet\textsuperscript{1}\\
  {\tt\small Jean.MARTINET@univ-cotedazur.fr}\and
 \hspace{100pt}I3S-CNRS/Universit\'e C\^ote d'Azur\textsuperscript{1} \quad $^2$INRIA/Universit\'e C\^ote d'Azur\textsuperscript{2}
  }

\maketitle

\begin{abstract}
In recent advancements in novel view synthesis, generalizable Neural Radiance Fields (NeRF) based methods applied to human subjects have shown remarkable results in generating novel views from few images.
However, this generalization ability cannot capture the underlying structural features of the skeleton shared across all instances.
Building upon this, we introduce \shortname: a novel generalizable human feature NeRF aimed at generating human biomechanic features using a pre-trained image encoder. While previous human NeRF methods have shown promising results in the generation of photorealistic virtual avatars, such methods lack underlying human structure or biomechanic features such as skeleton or joint information that are crucial for downstream applications including Augmented Reality (AR)/Virtual Reality (VR). \shortname~leverages 2D pre-trained foundation models toward learning human features in 3D using neural rendering, and then volume rendering towards generating 2D feature maps. We evaluate \shortname~ in the skeleton estimation task by predicting heatmaps as features. The proposed method is fully differentiable, allowing to successfully learn color, geometry, and human skeleton in a simultaneous manner. This paper presents preliminary results of \shortname , illustrating its potential in generating realistic virtual avatars with biomechanic features using NeRF.
\end{abstract}



\section{Introduction}
The development of custom virtual avatars capable of achieving photorealism is essential for realistic Augmented Reality (AR)/Virtual Reality (VR) environments. Moreover, it is a significant challenge to create a photorealistic virtual human avatar from a sparse set of images captured by a smartphone or a single camera. Previously, creating personalized virtual avatars with an underlying structure, such as a skeleton, required the use of costly camera setups that were only within the reach of a limited group of people. 
Furthermore, the labor-intensive process of body marker capture, extraction, and fitting of parametric models, such as SMPL~\cite{loper2023smpl}, is not scalable for widespread use.
\par The recent progress in Neural Radiance Fields has demonstrated significant potential in creating highly realistic virtual avatars using few images~\cite{hu2023sherf,su2021nerf}. However, previous NeRF-based methods do not provide any underlying structure, which is crucial for AR/VR applications and animation.
We introduce a novel approach named \longname, a unified framework to learn human biomechanic features such as the human skeleton with NeRF. Inspired by previous NeRF-based methods~\cite{yu2021pixelnerf, ye2023featurenerf} that utilize 2D encoders to generalize NeRF by conditioning input images or learning scene features. Our method uses a 2D pre-trained encoder to learn human features using NeRF architecture. 
Specifically, \shortnamewithspace predicts heatmap features of human joints, aiding in skeleton detection. Although this paper focuses primarily on skeleton detection, the architecture is adaptable to other biomechanic properties, such as body part segmentation.
Our method adopts two different types of encoder to generate features from images. \shortnamewithspace estimate separate heatmaps corresponding to each joint along with color and volume density. Our NeRF model takes as input the image feature of the 3D query point $\mathrm{x}$, along with its frequency encoding and view direction. The final heatmaps are generated using volume rendering inspired by the pixel color generation process of NeRF. 
\par In this paper, we present the initial results of \shortnamewithspace obtained with the RenderPeople dataset by distilling a state-of-the-art pose estimation algorithm based on heatmaps. To the best of our knowledge, our method is the first to estimate human biomechanic features with NeRF. Our contributions are as follows:
\begin{itemize}
    \item We present a new method for estimating human biomechanic features using NeRF. 
    \item We show that our model successfully learns to predict skeleton information from 2D images. 
\end{itemize}
\begin{figure}[t]
\begin{center}
   \includegraphics[width=1\linewidth]{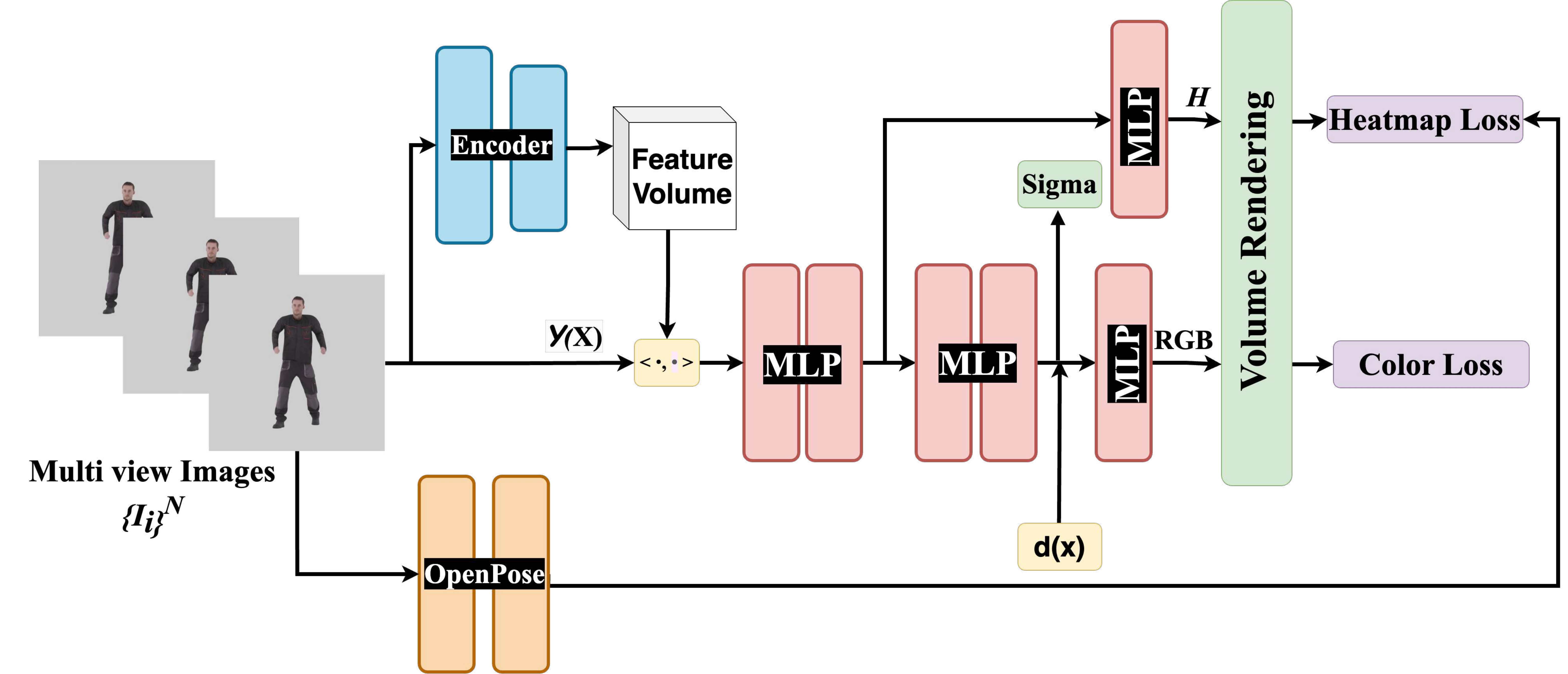}
\end{center}
\vspace{-10pt}
   \caption{\small Proposed pipeline of \shortname. }
   \vspace{-15pt}
\label{fig:method}
\end{figure}
\section{Method}
This section introduces \shortname, a unified framework utilizing the NeRF architecture for learning human features. It begins with a brief overview of NeRF, followed by the methodology for feature extraction using a 2D encoder, and concludes with a detailed description of skeleton extraction from heatmaps.\\
\textbf{Neural Radiance Fields:} The NeRF algorithm utilizes a multilayer perceptron (MLP) to map the 3D coordinate $\mathrm{x}=(x,y,z)$ and view direction $\mathrm{d}=(\theta, \phi)$ to the corresponding color $c$ and volume density $\sigma$. This mapping can be expressed as $F(\mathrm{x},\mathrm{d})\rightarrow(c,\sigma)$. Subsequently, volume rendering is employed to generate the pixel color by performing alpha composition of the volume density $\sigma$ and color $c$ using samples taken along the ray.\\
\textbf{Feature Extraction:}
We propose a novel architecture to estimate human features using the NeRF framework. NeRF models estimate the color $c$ and density $\sigma$ using an MLP, where the input is the positional encoding $\gamma(\mathrm{x})$ of the query point $\mathrm{x}=(x,y,z)$. 
The human feature $f(\mathrm{x})$, generated by the encoder corresponding to the input point $\mathrm{x}$, is concatenated with the positional encoding $\gamma(\mathrm{x})$ before being fed into the MLP.  In this work, we experimented with 2 different kinds of encoder, namely: ResNet~\cite{he2016deep} and DINO~\cite{caron2021emerging}.\\ 
\textbf{Learning Human Biomechanics features}
Previous methods\cite{yu2021pixelnerf} used encoded features to generalize NeRF, producing color and density as output. 
In this work, we extend the NeRF architecture to estimate human features by generating heatmaps of skeleton joints, as shown in Figure~\ref{fig:method}. We used an MLP with a skip connection, where the view direction is incorporated into the final layer before producing the color output. To generate heatmaps, we extract NeRF features from an intermediate layer and process them through a smaller secondary MLP.
We employ volume rendering to produce the final pixel color and heatmap values. This method is fully differentiable and optimized using a combined loss function: $l = l_c + \lambda_h l_h$ where $\lambda_h$ is the weighting factor.\\
\textbf{Skeleton prediction:}
We estimate the human skeleton by predicting joint locations from heatmaps. For each heatmap channel, which corresponds to a specific joint, a binary mask is generated through Gaussian filtering and thresholding. The joint locations are then identified as the pixels with the peak heatmap values within these mask regions.

\begin{table}[b!]
\vspace{-15pt}
    \centering
    \scalebox{1}{
    \begin{tabular}{c|ccccc}
    \hline
        Dataset & PSNR$\uparrow$ & SSIM$\uparrow$ & LPIPS$\downarrow$ & MSE$\downarrow$ \\
    \hline
        RenderPeople+ResNet & 46.421 & 0.9996  & 0.0024 & 0.0003 \\
        RenderPeople+DINO & 35.928 &  0.9914 & 0.0345 &  0.0001\\
    \hline
    \end{tabular}
    }
    \caption{\small Quantitative results on RenderPeople dataset.}
    \label{tab:res}
\end{table}
\begin{figure}[ht!]
\begin{center}
   \includegraphics[width=0.9\linewidth]{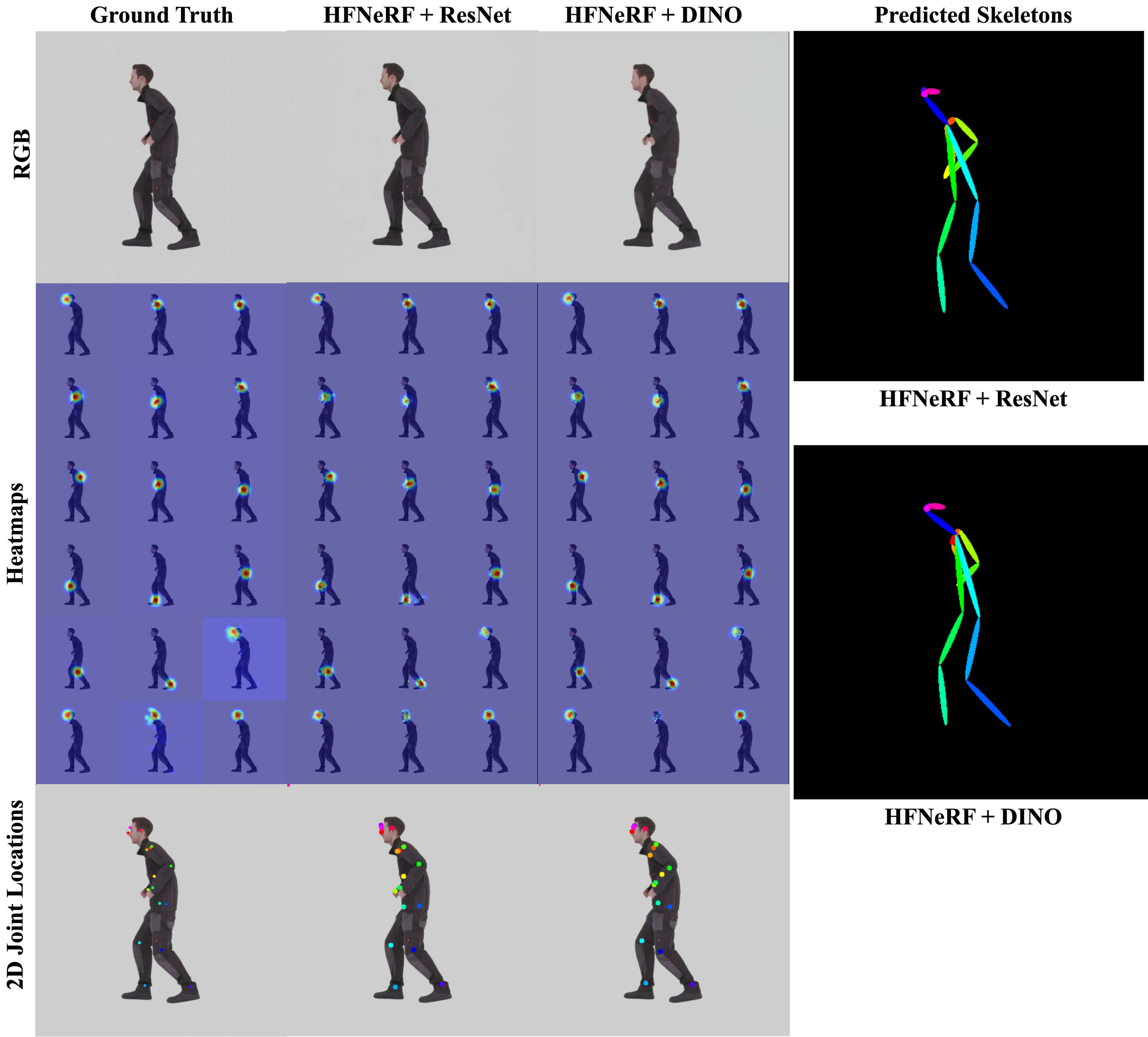}
\end{center}
\vspace{-10pt}
   \caption{\small Qualitative comparison on RenderPeople dataset.}
   \vspace{-15pt}
\label{fig:results}
\end{figure}
\section{Experiments and Discussions}
In this section, we present our preliminary results, focusing on skeleton detection and novel view synthesis. \\
\textbf{Dataset.} We trained our model on the RenderPeople~\cite{hu2023sherf} dataset, consists of multi-view image sequences of animated characters performing various actions. We use 34 cameras for training and 2 cameras for testing.\\
\textbf{Experimental setup.}
All experiments were conducted using a PyTorch implementation on an RTX 3090 GPU. For our experiments, the value of $\lambda_h$ was set to 0.5. We used the Adam optimizer for 100,000 iterations. We learn the heatmap features by distilling OpenPose~\cite{8765346}.\\
\textbf{Results.} This section details the initial results obtained using our \shortnamewithspace method. Quantitative results from the RenderPeople dataset are summarized in Table~\ref{tab:res}. The predicted and ground truth heatmaps are compared with the Mean Squared Error (MSE). The results indicate that ResNet features improve visual quality, while Vision Transformer-based DINO features lead to better heatmap predictions. Figure~\ref{fig:results} visually demonstrates these findings. In the future, we intend to expand our experimentation to encompass various datasets and perform additional comparisons with other methods.

\section{Conclusion}
This paper presents a novel framework called \shortname, which uses NeRF to learn human biomechanic features. Our initial findings demonstrate the effectiveness of \shortnamewithspace in predicting human features, a significant improvement over previous NeRF methods for humans. Although our focus was on human skeleton detection, we believe that this architecture can be extended to other generalizable human features, such as body part detection.
\section{Acknowledgements}
This project has received funding from the H2020 COFUND
program BoostUrCareer under MSCA no.847581.

\bibliographystyle{abbrv}
\bibliography{biblio}


\end{document}